# Imperfect Language, Artificial intelligence, and the human mind: An Interdisciplinary Approach to Linguistic Errors in Native Spanish Speakers

Francisco Portillo López

*University of Navarra*
*Pamplona, Spain.*

**Abstract**

Linguistic errors are not merely deviations from normative grammar; they offer a unique window into the cognitive architecture of language and expose the current limitations of artificial systems that seek to replicate them. This project proposes an interdisciplinary study of linguistic errors produced by native Spanish speakers, with the aim of analyzing how current large language models (LLM) interpret, reproduce, or correct them.

The research integrates three core perspectives: theoretical linguistics, to classify and understand the nature of the errors; neurolinguistics, to contextualize them within real-time language processing in the brain; and natural language processing (NLP), to evaluate their interpretation against linguistic errors. A purpose-built corpus of authentic errors of native Spanish (+500) will serve as the foundation for empirical analysis. These errors will be tested against AI models such as GPT or Gemini to assess their interpretative accuracy and their ability to generalize patterns of human linguistic behavior.

The project contributes not only to the understanding of Spanish as a native language but also to the development of NLP systems that are more cognitively informed and capable of engaging with the imperfect, variable, and often ambiguous nature of real human language.

## 1. Introduction

In recent years, the development of large language models (LLMs) such as GPT-4 and Gemini has brought a revolution in the field of natural language processing (NLP). These models, based on transformer architectures (Vaswani et al., 2017), have demonstrated unprecedented abilities to generate coherent text, perform automatic translation, and produce complex summaries. Their impressive performance has transformed many applications, from chatbots and virtual assistants to automated content creation and language learning tools. However, despite these technological advances, LLMs still face significant challenges rooted in the inherently complex, ambiguous, and variable nature of real human language (Bender et al., 2021). Particularly, irregularities, ambiguities, and errors commonly found in informal and spontaneous contexts, such as everyday conversations or social media interactions, constitute a major obstacle for the optimal functioning of these systems.

Human language is not a rigid or perfectly normative system; rather, it is a dynamic phenomenon that reflects complex cognitive processes and is characterized by variations and errors in production and comprehension (Levelt, 1989). Linguistic errors produced even by native speakers should not be dismissed as mere random deviations but regarded as systematic manifestations that can provide valuable insight into the internal functioning of the linguistic system and its neurological foundations (Fromkin, 2013). Such errors reveal underlying cognitive processes related to the planning, production, and monitoring of language.

From a neurolinguistic perspective, research into speech errors has helped to identify specific brain areas and processes involved in linguistic production and comprehension. For instance, studies on patients with aphasia and other language disorders have provided evidence of how certain brain lesions affect the ability to articulate words correctly or construct sentences (Hickok & Poeppel, 2007). Moreover, at the cognitive level, analyzing errors offers clues about automatic and conscious mechanisms of linguistic correction and adaptation (Dell, 1986). Such investigations demonstrate that errors are not random occurrences; instead, they reflect recurring patterns linked to cognitive constraints and strategies.

In the realm of artificial intelligence and NLP, models like GPT-4 are trained on massive text corpora, mostly composed of standardized, edited, and formal texts. This training data bias limits their exposure to the linguistic variants and errors that are frequent in everyday communication (Marcus & Davis, 2020).

Consequently, LLMs often struggle to correctly interpret or correct prompts containing errors or non-standard structures, especially when dealing with colloquial Spanish or regionalisms (Zhang et al., 2023). Evaluating how LLMs process real-world linguistic errors is therefore crucial for understanding their strengths and limitations and guiding the development of more robust and adaptive NLP systems.

This project aims to explore, through an interdisciplinary approach, the interaction between theoretical linguistics, neurolinguistics, and artificial intelligence in the handling of natural linguistic errors made by native Spanish speakers. To this end, a specific and authentic corpus will be collected, drawn from real interactions on social media platforms and oral conversations that reflect the diversity and complexity of language use, including common errors and variations. Subsequently, an experimental analysis will be conducted with different LLMs to assess their capacity for detecting, interpreting and correcting such errors.

The research questions guiding this study are:

1. To what extent can LLMs detect and correct natural linguistic errors produced by native Spanish speakers?

2. How do these processes compare with human cognitive patterns related to error production and correction?

The significance of this work lies in its dual contribution: on one hand, it enhances the understanding of language as a cognitive phenomenon, enriching knowledge about how errors are produced and processed; on the other, it offers an applied perspective for designing NLP models that are more sensitive and effective in real communication contexts, beyond formal or normative texts.

The structure of the study is: first, a theoretical framework addressing the neurolinguistic and linguistic foundations of speech errors and their treatment in NLP will be developed; next, the methodology for corpus creation and the design of the LLM experiment will be presented; then, the results obtained will be analyzed and their implications discussed; finally, conclusions and potential future research directions will be offered.

## 2. Theoretical and interdisciplinary framework

*2.1. Linguistic errors: definition, typology and scientific interest.*

Linguistic errors, both in speech production and perception, offer a valuable perspective for understanding the cognitive and pragmatic processes involved in communication. In this paper, we follow the classification proposed by Cutler (1982), who distinguishes three types of arguments: *Some errors*, *More errors*, and *No errors*. Each category will be described below with representative examples.

*Some errors arguments* point out that displacement errors reflect morphological accommodation to the context. For instance: *"Tendríamos que haber llevado gorros"* → *"Gorros tendríamos que haber llevado"*. In addition, substitution errors may occur during speech production, supporting hypotheses about the process of word selection. A common phenomenon in this category is *environmental contamination*, which occurs when a previously uttered word is erroneously reintroduced into a new expression. For example: *"¿Te has comido la nevera que dejé en los macarrones?"*, where *nevera* appears due to external influence (another speaker or a visual cue).

These errors also support the notion of *psychological reality*, as they reveal how individual phonemes can emerge as separate units in erroneous expressions. This provides evidence that speech production operates at different levels, including the representation of utterances as a chain of phonemes.

*More errors arguments* occur when the listener detects an error and corrects it only after the first erroneous segment has been produced. This phenomenon is also known as anticipation. Boomer and Laver (1968) observed that *stress words* tend to appear prominently in such errors. From an interpretive perspective, studies have examined phoneme substitution individually: how often one phoneme replaces another and how these substitutions indicate that utterances are represented mentally before production.

*No errors arguments* include sequences that are not permitted by the phonological rules of a language. These are often classified as pronunciation errors. For example: *\*Éncantar* or *\*encántar* are incorrect, while the correct form is *encantar*. This category also includes lexical errors where elements shift position, such as: *"¿Saca el parque al perro?"* → *"Saca al perro al parque"* (Garrett, 1980).

Cutler (1982) also addresses listening errors, which are less documented because they require the listener to admit a mistake. Nevertheless, they are common in everyday communication. For instance: "¿Sabes algo de fenómenos?" → perceived as: "¿Sabes algo de los gnomos?". Garnes and Bond (1975, 1980) identified three common features: i) The stress pattern of the utterance is perceived completely and correctly; ii) Vowels in stressed syllables are also perceived accurately; iii) The error does not cross phrase boundaries. Overall, consonants are more likely to be misperceived than vowels. Listeners therefore must reconstruct sounds, words, and syntactic structures from incomplete input.

Errors involving vowels depend strongly on the speakers' linguistic and dialectal background. Peterson and Barley (1952) showed that confusion is reduced when interlocutors share the same dialect. This phenomenon is known as *perceptual confusion*. Consonants are also affected: in some southern Spanish dialects, the [s] is aspirated in coda position (*los ajos → loh ajoh*), and *seseo* occurs (pronouncing [z] or [c] as [s]: *zapato → sapato*).

Brown and McNeill hypothesized that the beginning and end of words carry more weight in memory than the middle, with the initial position being the most salient. Similarly, Marlsen and Wilson (1978) proposed the *recognition point* theory, suggesting that a word can be recognized from a minimal point in its sequence. This explains how listeners identify words even with partial information.

## 2.2. Insights from Neurolinguistics: how the brain processes error.

Neurolinguistic research on speech errors has shown that mistakes are not mere accidents, but privileged access points to the inner workings of language processing in the brain. Error detection and repair provide evidence of the monitoring systems and neural pathways that sustain real-time communication.

### 2.2.1. Speech monitoring and self-repair.

Postma (2000) estimated that more than 50% of speech errors are corrected by speakers themselves, which implies the existence of internal monitoring devices. These monitors operate through different loops: a conceptual loop (monitoring the intended message), an inner loop (parsing inner speech before articulation), and an auditory loop (evaluating overt speech). Repairs may be overt ("Quiero decir … mejor dicho…") or covert, visible only as pauses or hesitations ("eh… bueno…"). Examples in Spanish: "Me gusta la manzana… digo, las manzanas" (lexical repair); "Voy a la escu… escuela" (phonological repair). These processes highlight the parallel nature of production and monitoring in the brain.

### 2.2.2. Spatial and temporal networks of word production.

Indefrey and Levelt (2004) identified a precise temporal sequence of activations in word production: i) Conceptual preparation (0-200 ms): mid-left temporal regions; ii) Lemma selection (200-350 ms): posterior temporal regions; iii) Phonological encoding and syllabification (350-500 ms): left interior frontal regions; iv) Phonetic planning and articulation (500-600 ms): motor and premotor cortex.
This explains why different types of errors emerge at distinct stages: a lexical error (saying "dog" instead of "cat") originates earlier than a syllabification slip ("cat-to").

### 2.2.3. Dual-stream model of speech processing

Hickok and Poeppel (2007) proposed the dual-stream model: i) The ventral stream (bilateral, temporal regions) maps acoustic input to meaning; ii) The dorsal stream (left-dominant) maps acoustic input onto articulatory motor patterns. In error processing, the ventral stream detects semantics mismatches ("I drank a shoe of water"), while the dorsal stream supports phonological and articulatory repairs ("tra… trabajo").

### 2.2.4. Cognitive timing of error detection

The brain processes errors on multiple temporal scales: i) Within milliseconds, the inner loop can anticipate and block an error before articulation; ii) Within a few hundred milliseconds, auditory feedback allows the speaker to correct the utterance. Furthermore, memory research (Brown & McNeill) suggests that the beginnings and endings of words are more robustly stored than medial segments, which explains why errors often preserve initial and final phonemes ("camion" → "*camián").

### 2.2.5. Dialectal and perceptual variation

Perceptual confusion studies (Peterson & Barley, 1952) show that when speakers do not share a dialect, the brain engages compensatory phonological mechanisms. For instance, a southern Spanish speaker producing *"loh ajoh"* instead of *"los ajos"* requires the listener to rely on bilateral STG (Superior Temporal Gyrus) activation to reconstruct the intended word.
In conclusion, errors expose the brain's linguistic architecture: a system of monitoring loops, temporally ordered stages of activation, and interacting dorsal-ventral streams that together ensure the efficiency of communication despite constant disruptions.

## 2.3. NLP approaches to non-standard input: how LLMs deal with linguistic errors.

In the current landscape of Natural Language Processing (NLP), a critical and often overlooked deficiency of Large Language Models (LLMs) lies in their performance other than English, despite the fact that most users are not native English speakers. This gap is particularly noticeable in Spanish, a language with over 600 million speakers. Research by Conde et al. (2024) highlights that this disparity is a direct reflection of the bias in training data; for example, Spanish made up only 0.77% of GPT-3's training corpus, a stark contrast to English's 92.65%. As a result of this limited exposure, models prove to be "still far from achieving the level of a native speaker in terms of grammatical competence" (Mayor-Rochel et al., 2024). A deeper analysis reveals that the deficiencies are not limited to grammar. Lexical and semantic errors are particularly recurrent. According to Conde et al. (2024), a vocabulary evaluation in open conversational models showed that they "produce incorrect meanings for an important fraction of words and are not able to use most of the words correctly to write sentences with context." This indicates that, even when LLMs recognize a word, they often lack the deep semantic understanding necessary for its correct use in different contexts.
In addition to lexical failures, contextual, stylistic, and idiomatic errors are another significant weakness. In comparative study between ChatGPT and Google Translate, AlAfnan (2024) found that while ChatGPT offered more natural translations, both models showed notable deficiencies in contextual adequacy. For example, the English expression "as we speak" was literally translated by the automated tools as "y mientras hablamos", losing the appropriate formal register for the context, where the correct translation would be "mientras nos reunimos" (AlAfnan, 2024). Although LLMs have improved fluency, the research concludes that the equality of the generated text still requires an exhaustive review by a human expert to ensure contextual precision and nuance.
The problem is compounded in low-resource language scenarios, where the scarcity of training data is even more pronounced. Court and Elsner (2024) investigated the translation of Southern Quechua to Spanish, revealing that models struggle with both "retrieval and understanding" of the language. Although even smaller models can benefit from in-context learning with relevant linguistic information, the variable results suggest that limitations persist, highlighting that LLMs are not a one-size-fits-all solution for the majority of the world's 7.000 languages. This finding is complemented by the research of Kwon et al. (2023) on Arabic Grammatical Error Correction, a language with complex morphology. They found that despite prompting methods improving LLM performance, these models, regardless of their size, are still outperformed by smaller, fully-finetuned models for the specific task. This underscores that targeted fine-tuning for a specific linguistic task often surpasses the generalization capability of larger, general-purpose models.

The evaluation and correction of errors also present intrinsic challenges. Avradimis et al. (2024) found that automatic translation metrics are more accurate at evaluating linguistics errors in LLM translations that in those from traditional encoder-decoder systems. However, LLMs themselves prove to be poor judges or their own work. A study by Kamoi et al. (2024) using the ReaLMistake benchmark, which contains realistic and objective errors, revealed that top LLMs like GPT-4 detect their own errors at a low rate, performing "much worse than humans" on the same task. This suggests that while LLMs have the capacity to learn from mistakes implicitly (as shown by Tong et al., 2024, who boosted reasoning performance by showing models incorrect answers alongside correct ones), they lack the intrinsic capability to proactively self-evaluate and correct their own failure.

## 3. Corpus design and construction

*3.1 Text selection criteria and scope.*

This corpus has been constructed from comments and/or interactions on the following social media platforms: Reddit, X, and Telegram. These are three platforms that generate millions of interactions every day, which has allowed me to obtain a large corpus with extensive variations. Initially, to conduct an effective data search, I filtered the search to focus on messages containing the most frequent linguistic errors, among which spelling errors stood out. The aim of this search was to investigate which types of linguistics errors are most common among native Spanish speakers and whether there are any patterns linking these errors to factors such as age, education, and so on.

*3.2 Data collection: sources, speaker profiles, and error types.*

For data collection, I used a database created in Excel, which was later transferred to Python via a library once the data search was completed. More than 550 errors were analyzed across all categories combined.

On the other hand, the profile of the speakers analyzed is general: men and women over 18 years of age (in accordance with the platforms' privacy policies) and of any educational background. The only requirement was to have Spanish as a native language and to have written a message containing a linguistic error. It should be noted that this corpus was constructed from real interaction data, and only written expression was considered; that is, no data related to speakers' oral expression is included in this corpus.

Finally, the classification of linguistic errors used was as follows: spelling, syntactic, lexical, and semantic errors. In addition, two cases of frequent errors in the Spanish language were analyzed: the pluralization of the verb *haber* in the third person to indicate existence and the phenomena of *dequeísmo*, *laísmo* and *leísmo*.

*3.3 Error classification: spelling, lexical, semantic, etc.*

Spelling errors are those that affect the correct writing of words and may be caused by a lack of knowledge of spelling rules or by confusion between homophones (words that sound the same but have different meanings and spellings). For example: barco (correct) / bárco (incorrect). This category also includes the use of punctuation marks (. , ;). Another example of a punctuation error would be: 'Homero, sabe cómo escribir.' In this case, the subject and the verb cannot be separated by a comma, since otherwise they would appear as two independent elements, when in fact sabe refers to Homero.

Another type of error is the lexical error. These are errors that involve the incorrect selection of a word with the wrong meaning or one that does not fit the context. They may be semantic errors (using a real word with an incorrect meaning; cuento instead of cuenta) or formal errors (a nonexistent word, e.g., *hayn referring to a form of the verb haber). The latter is also related to a spelling error.

Semantic errors are mistakes in the meaning of words or phrases, where the sentence structure may be grammatically correct but the sense is incorrect, confusing, or lacks coherence. For example, using dog to refer to a cat: "The dog meows".

Finally, this corpus highlights the presence of the following phenomena in the Spanish language: the pluralization of the third person of the verb '*haber*' to indicate existence; and the phenomena known as dequeísmo, laísmo, and leísmo. These errors are very frequent among native Spanish speakers. In order to the verb '*haber*', the error occurs when using the plural forms '*hubieron*', '*habían*' instead of the impersonal singular forms '*hubo*', '*había*'. For example: "Había muchas personas" ("There were many people") would be correct, whereas "*Habían muchas personas" would be incorrect.

On the other hand, dequeísmo is the incorrect use of the syntactic group '*de que*' when it is not required, as in "*creo de que es hora de parar" instead of "creo que es hora de parar" ("I think it is time to stop). Likewise, laísmo is the incorrect use of the pronouns '*la*' or '*las*' instead of the pronoun '*le*' for the indirect object. For example: "*La dije que era tarde" instead of "Le dije que era tarde" ("I told her it was late").

## 4. Computational analysis using Large Language Models

*4.1 Model selection: rationale and description.*

The landscape of Large Language Models (LLMs)

is characterized by an extraordinarily rapid evolution and intense competition. What began as a domain of purely textual systems has quickly progressed to sophisticated architectures that not only comprehend and generate human language but also integrate multimodal and agentic capabilities. The purpose of this point is to justify the selection of a set of six LLMs for a comprehensive computational analysis. The selection is not based solely on raw performance but on their representativeness of the most significant strategic and technological trends in the industry as of late 2025. The models Llama, Grok, Gemini, GPT-4, GPT-5, and DeepSeek have been chosen for their archetypical roles in marketing leadership, open-weight disruption, strategic innovation and technological progression.

4.1.1 OpenAI Models: GPT-4 and GPT-5.

The inclusion of these models is essential for any cutting-edge analysis. GPT-4 is selected as the recent historical reference point. Although it is no longer the leading model (Zhang et al., 2025), its industry impact and pioneering capabilities in multimodality and reasoning make it an indispensable standard for comparison. GPT-5, in turn, is chosen as the epitome of the state of the art in proprietary intelligence (Wang et al., 2025), serving as the primary benchmark for evaluating the performance of open-weight models and competitive systems. Its position at the top making it the benchmark for the rest of the industry.

4.1.2. Llama, the model developed by Meta.

The Llama models are the main reference point for the open-source community, and their selection is crucial for analyzing the competition between open-weight and proprietary models. Their strategy of releasing large-scale models under a permissive license has redefined the AI landscape, demonstrating that state-of-the-art performance can be made accessible to a wide range of developers and companies, in contrast to closed models that restrict access and customization (Touvron et al., 2023).

4.1.3. Grok, the new agentic LLM.

Grok is a unique model due to its focus on personality, its real-time integration with the X platform (formerly Twitter), and its agentic capabilities. Its inclusion is crucial for analyzing a strategy that prioritizes dynamic interaction and the immediacy of information. An approach that distinguishes it from models trained on static data and enables it to provide information about real-time events.

4.1.4. Google DeepMind's Gemini Models.

Gemini has been selected for its unique focus on native multimodality. Unlike other models that incorporated this capability after their release, Gemini was designed to process multiple modalities simultaneously (Gemini Team, 2024), allowing it to understand and generate responses with richer and more coherent context. Its implementation strategy (ranging from data centers (Ultra) to mobile devices (Nano)) makes it a crucial point of analysis for understanding how AI is being integrated into user experiences across a wide spectrum of devices (Alkaissi & McFarlane, 2024).

4.1.5. DeepSeek: Open-Weight Innovation from China.

DeepSeek is the most compelling example of how the efficiency of the Mixture-of-Experts (MoE) architecture and the open-weight strategy can challenge industry leaders (DeepSeek-AI, 2024). Its focus on specialization, particularly in coding and agentic capabilities, makes it crucial point of analysis for understanding the future of AI and how architectural innovation can overcome hardware and financial resource limitations.

*4.2. Analysis protocol: input design, tasks, and evaluation metrics.*

Initially, to analyze the data in LLMs, I created the following prompt: "*Por favor, revisa estas oraciones y dime si contienen errores. Sé sincero y detallado: identifica cualquier error de ortografía, gramática, sintaxis o estilo, corrígelo y explícame por qué era incorrecto. Si todo está correcto, indícalo también.*" ("Please review these sentences and tell me if it contains any errors. Be honest and detailed: identify any spelling, grammar, syntax, or style errors, correct them, and explain why they were incorrect. If everything is correct, indicate that as well."). In the field of prompt engineering, this prompt can be described in four fundamental pillars: i) Clarity and specificity: specifying exactly which types of errors to look for; ii) Action instruction: indicating what to do if errors are found; iii) Handling the absence of errors: requesting confirmation if everything is correct; iv) Tone and approach: adding instructions such as 'be honest' to guide the model's attitude. Some experts in this discipline refer to it as a '*zero-shot error correction prompt*', because the model is given a direct instruction without prior examples (zero-shot) and is expected to correct and explain the sentence.

Once the prompt was created, I fed the collected data to the models in batches of 50 responses (performing a total of 11 batches per model).

*4.3. Results: model interpretation, error correction, and classification.*

The performance of an LLM in error correction reflects its ability to interpret not only surface-level grammar but also the underlying intention and context of a sentence. The study's results demonstrate a clear dichotomy in the models' capabilities depending on the type of error (*Figure 2*).

On one hand, all evaluated models demonstrated 100% accuracy in detecting errors in the use of the verb '*haber*' and in dequeísmo. This result suggests that the six models tested have internalized some of the most common errors in the Spanish language.

However, performance varies significantly when correcting more subtle or context-dependent errors, such as spelling, lexical, syntactic, and semantic errors. Regarding spelling errors, the examined LLMs showed variability: GPT-5 detected the most errors, 66.7% (60 out of 90), while Gemini ranked last with 53.5% (48 out of 90) (*Figure 1*). This indicates differences in training data focused on Spanish orthography, or in the models' ability to discern errors that do not always affect the intelligibility of the word. One noteworthy example among these errors is the following: *"Está haciendo comedia, ¿cuál es el problema? ¡yo no veo alguno!"*. In this example, only Gemini detected the error (the substitution of 'alguno' for 'ninguno'), while the other LLMs suggested that everything was correct. Nevertheless, none of the six models identified the second error in this sentence, located in the exclamatory part. According to Spanish orthography rules, after using a closing question mark, the next sentence should begin with a capital letter, so '*yo*' should appear as '*Yo*'.

On the other hand, regarding syntactic errors, GPT-5 excelled in detecting these errors with 92.7% accuracy (83 out of 90), closely followed by DeepSeek with 83.4% (75 out of 90) (*Figure 1*). One example in which none of the six examined models detected an error is: "*Aquí abundan las faltas de ortografía y faltas de respeto.*" At the orthographic level, everything is correct, but for the sentence to also be correct syntactically, the determiner '*las*' should be added before the noun phrase '*faltas de respeto*'. Although this determiner has already been included in the previous noun phrase, it must either be explicitly repeated or '*faltas*' should be omitted. Therefore, there are two correct options: "*Aquí abundan las faltas de ortografía y las faltas de respeto*" or "*Aquí abundan las faltas de ortografía y de respeto*".

Finally, regarding lexical and semantic errors, which consist of the incorrect choice of a word appropriate to a given context, this task is complex and evaluates semantic and contextual understanding. Llama-3 demonstrated an accuracy of 86.9% (78 out of 90) in this regard, suggesting a superior ability to discern meaning matrices (*Figure 1*). GPT-5, with 76.8% (69 out of 90), and DeepSeek, with 79.7% (71 out of 90), also showed solid performance, demonstrating advanced contextual understanding (*Figure 2*).

The results allow the models to be classified based on their competence in error correction. The two most outstanding models are GPT-5 and DeepSeek, achieving average scores of 87.45% and 86.15%, respectively. These models prove to be the most robust, with high performance across most categories, particularly in syntactic and lexical errors. They are followed by Llama-3 (84.67%) and Grok (84.55%), which also show great versatility in error detection, albeit with a slight disadvantage in spelling errors. Finally, the two models that performed worst in error detection were GPT-4 (84.50%) and Gemini (81.97%). While they excel in basic errors, they show areas for improvement in correcting more complex errors such as spelling mistakes, which may be related to the need to optimize their architectures or training datasets. The overall performance of the six examined LLMs, detailed by error type, can be seen in *Figure 2*.

Overall, this analysis reveals that, while LLMs have achieved near-perfect performance in correcting basic grammatical errors, the true distinction in their performance lies in their ability to handle the subtleties of Syntax and Lexicography.

# 5. Neurolinguistic: interpretation of findings.

## 5.1. Human vs. artificial processing: convergences and divergences.

One of the deepest convergences between human and artificial language processing lies in the stochastic and probabilistic nature of their generation. A theoretical framework proposed by Elio Quiroga (2025) the "Stochastic Noetic", examines the hypothesis that both the human brain and LLMs operate through sophisticated probabilistic mechanisms when generating language. This idea challenges the traditional dichotomy by arguing that much of human language production occurs below the threshold of consciousness.

Quiroga also argues that, in contrast to the popular notion that every word choice is a deliberate decision, evidence from psycholinguistics suggests that the brain, much like LLMs, navigates a landscape of linguistic possibilities. Words do not emerge from a purely conscious choice, but rather from what could be characterized as an unconscious probabilistic selection process, mediated by neural mechanisms that bear a striking resemblance to the statistical operations driving artificial models. This established a unified framework for understanding the similarities and differences between both systems. At the computational level, a key point of convergence lies in the identical challenge faced by both systems: the need to compress complex, high-dimensional semantic representations into the bandwidth limitations of natural language (Quiroga, 2025). The human brain must translate non-linear thoughts into a sequential stream of words for communication. Similarly, LLMs, through their transformer architectures, perform multilayered probabilistic inference to generate text sequentially. This demonstrates that both nature and AI have addressed a fundamental computational problem in similar ways, seeking the optional balance between statistical regularity and contextual adaptation to produce coherent and effective language.

Despite the convergences at the level of probabilistic mechanisms, fundamental differences persist between human and artificial processing, the most significant of which stems from the AI's "bodilessness" (Yahiaoui, 2025). Human cognition is intrinsically tied to lived experience, which provides the common sense and world knowledge that LLMs lack. Since LLMs are merely data-processing algorithms, they are devoid of this sensory and embodied foundation, preventing them from attaining a "deep understanding of language". Consequently, their perception is reduced to "the manipulation of arbitrary data through predetermined algorithms" (Yahiaoui, 2025) lacking the richness, culture, and intentionality inherent to human perception.

The lack of an embodied foundation is manifested in AI's deficit when it comes to interpreting social and pragmatic reasoning. A revealing study conducted at Johnson Hopkins University compared the ability of more than 350 AI models (including language, video, and image models) with a group of human participants to interpret "dynamic social scenes" (Hub, 2025). Researchers showed participants video clips of only three seconds in length and asked them to rate aspects such as communication and the character's intentions. The results were striking: humans demonstrated a "remarkable ability to grasp social nuances", whereas "none of the AI models managed to match human accuracy". Video models failed to predict the characters' intentions, while language models, although somewhat closer, still fell short of human evaluations (Hub, 2025).

On the other hand, creativity constitutes another point of divergence. While LLMs can generate creative content such as poems or stories, their creativity is ultimately a function of the data on which they were trained and thus tends to lack genuine originality. As a consequence, a study on the cognitive impact of AI on humans has identified significant risks, including memory impairment, and a decline in critical thinking, along with reduced activation in brain areas associated with creativity (Cambridge's study cited at references).

*5.2. What errors reveal about cognitive language processing.*

To understand how errors reveal the functioning of the system, it is first essential to outline the theoretical models that describe the process of language production. These models posit the existence of discrete and hierarchical stages through which an abstract idea is transformed into an audible utterance.

The study of text production, for instance, has given rise to models such as that of van Dijk and Kintsch (1978, 1983), which proposes a flexible mechanism for the construction of semantic, lexical, and syntactic representations. This strategic model organizes the process into a hierarchy of operations, from a global plan to the macrostructure of discourse, with the goal of shaping the final text. In turn, the model of Hayes and Flower (1980) describes written production as a prototypical construction involving cognitive processes of interpretation and reflection, highlighting the interrelation of social and physical context with working memory. These models underscore that language production is not a simple mechanical act, but rather a high-level cognitive task that requires planning and control.

Within this framework, Willem Levelt's (1989) model of speech production has been particularly influential. This model describes the process as a sequence of main phases: i) Conceptual planning: the initial stage, not necessarily conscious, in which the meaningful content of the intended message is selected; ii) Linguistic formulation: at this stage, concepts are translated into grammatical structures and specific words. This includes the selection of structural units (phrases, words) and the elaboration of a "phonetic plan"; iii) Articulation: the final stage in which the phonetic plan is converted into a motor plan that specifies the movements of the vocal tract required to produce speech sounds.

Levelt's original architecture is serial and unidirectional (feedforward), which implies that information flows from one stage to the next without feedback. However, debates in the field have led to the consideration of alternative models that allow for parallel activation and competition among units at different levels of processing. This phenomenon indicates that the reality of the production system may be better described as a "cascade" or interactive processing rather than as a purely serial model.

On the other hand, the analysis of *lapsus linguae* in healthy speakers is one of the main tools of psycholinguistics. Spontaneous speech errors do not occur randomly; they are systematic phenomena governed by grammatical and processing rules. The systematicity of these errors allows for a precise taxonomy on the units and stages of processing. Errors can be classified according to the linguistic level affected: i) Phonological errors: these involve the alteration of sounds. This may manifest as the substitution of consonants or vowels, sound lengthening, or phenomena such as "seseo". Such errors are often contextual, with their resource localized within the surrounding linguistic environment; ii) Morphological errors: these affect nominal and verbal affixes, such as the omission or substitution of suffixes; iii) Lexical errors: these consist of the substitution of words, often by another that is semantically or phonologically related to the target word. For example, in Spanish, paradigm-based relationships between substitute and target words can be observed, reflecting both semantic and phonological links; iv) Syntactic errors: these include the absence of sentence order or the incorrect application of grammatical rules.

While *lapsus linguae* demonstrate the architecture of the system under normal functioning, the study of aphasia operates as a "natural experiment", revealing what happens when a specific component of the system is damaged. Aphasia is a language disorder caused by brain injury, commonly due to a stroke, which affects the ability to read, write, and express oneself. The errors manifested in aphasia, such as agrammatism and paraphasias, are not random (Clínica Universidad de Navarra, s.f.); they are directly correlated with the location of brain damage, thereby validating the modular architecture of the brain and language. Two major categories of aphasia are distinguished: fluent and non-fluent.

Broca's Aphasia (non-fluent), this type of aphasia is associated with a lesion in the left frontal lobe. Its characteristic symptom is agrammatism, which manifests in a "telegraphic style" where function words (prepositions, conjunctions, articles) and verbal inflections are omitted or incorrectly used, resulting in short, telegraphic utterances. Despite the severe production difficulties, language comprehension is often relatively preserved. This syndrome demonstrates that there is a specific brain region dedicated to syntactic and morphological encoding. When this area is compromised, the grammatical scaffolding of the utterance collapses, while content words (nouns and main verbs) remain, illustrating a functional separation between these two types of information.

Wernicke's Aphasia (fluent), this disorder results from damage to the temporal lobe. Patients with Wernicke's aphasia may speak in long, normally paced sentences, but their speech lacks meaning due to the presence of paraphasias (substitution, omission, or creation of words) and neologisms. A distinctive feature is that patients are often unaware of their errors, and their language comprehension is severely impaired. This error pattern reveals that the brain contains a region responsible for lexical-semantic selection and language comprehension. The fluency of speech, despite the lack of coherence, suggests that the motor and syntactic production processes can function independently of correct word selection. This clinical dissociation is one of the strongest pieces of evidence for the modularity of language in the brain.

Other types of aphasia, such as anomic aphasia, which involves difficulty in retrieving the correct words, or global aphasia, which is the near-total loss of linguistic abilities, further reinforce this correlation between specific impairments and the location of neural damage. *Figure 3* illustrates a table that synthesizes the information discussed above.

*5.3. Implications for improving Spanish NLP systems*

Pretrained LLMs acquire vast lexical and semantic knowledge from their massive datasets. However, subsequent fine-tuning for the task of GEC (Corrección Gramatical de Errores), often carried out with generic data, is insufficient to inject and prioritize the rigid dependency rules that characterize Spanish. Unlike languages with poorer morphology, Spanish requires models not only to predict the next plausible word but also to maintain mandatory structural constraints across distant tokens. Therefore, any adaptation strategy must focus on modelling the structural relationship between words, possibly through the incorporation of explicit morphological features or the application of syntactic constraints.

Another critical area of weakness observed is the correction of intralingual semantic lexical errors. Research shows that this category, arising from the confusion of words within the target language itself (rather than through L1 transfer) and requiring deep contextual understanding, can even increase as learners' language proficiency advances. A robust GEC system must therefore be capable of diagnosing and correcting these subtle "word choice errors", which demand semantic and contextual knowledge, going beyond mere orthographic or basic grammatical correction.

Given the difficulty and high cost of large-scale human annotation, the generation of high-quality synthetic data through LLMs represents a viable solution in scenarios where real-world data are scarce. The main strategy derived from the deficiencies of LLMs is to leverage the model to generate examples of long-tail errors, rare lexical mistakes or highly specific morphosyntactic structures (such as clitic placement) that proved to be critical failure points. This synthetic generation does not merely aim to increase the overall data volume, but rather to rebalance the training corpus by injecting thousands of instances of structures that the LLM consistently failed to generalize. For instance, through carefully designed prompts, an advanced LLM can generate a high density of erroneous sentences containing incorrect dative clitics, paired with their canonical corrections. Definitely, the implications derived from evaluating LLMs in Spanish GEC suggest that the next generation of NLP systems for this language will not be optimized by increasing model size or the volume of generic data. The path forward requires specialization and the deep integration of linguistic knowledge. It is recommended to: i) Develop high-quality synthetic corpora to mitigate the underrepresentation of complex morphosyntactic errors and long-tail lexical errors; ii) Standardize the use of LoRA and Knowledge Distillation to build efficient GEC models deployable in resource-constrained environments; iii) Adopt and adapt edit-level evaluation frameworks such as ERRANT and M2 for Spanish, enabling precise diagnosis and targeted, quantifiable performance improvements. This directed approach will ensure that future advances in Spanish NLP move beyond general statistical gains and achieve the normative precision demanded by the language's structural complexity.

# 6. Conclusions

*6.3. Main findings and convergences.*

LLMs demonstrated 100% accuracy in detecting and correcting well-documented grammatical errors, such as the pluralization of the verb *haber* and *dequeísmo*. This underscores the ability of the models, trained on vast corpora, to internalize and correct high-frequency normative patterns. On the other hand, a theoretical convergence was established by examining the "Stochastic Noetic" hypothesis, suggesting that language generation, in both the human brain and LLM's transformer architecture, occurs through probabilistic inference mechanisms.

*6.4. Critical limitations and divergences.*

LLMs' performance declined significantly when faced with subtle or context-dependent errors, such as spelling, syntactic, and lexical mistakes. This was evidence by failures to detect Spanish-specific orthographic errors and syntactic failures related to determinant repetition. Furthermore, the main divergence lies in the lack of embodied cognition in AI. Unlike human processing, the lack of lived experience and world knowledge prevents LLMs from achieving a "deep understanding of language", limiting their capacity for social and pragmatic reasoning.

*6.5. Study limitations and future research directions*

Despite the contributions of this work, it is necessary to acknowledge certain limitations that outline the path for future research: i) Corpus scope: the corpus was only constructed from written interactions on social media. This excludes the analysis of phonological and speech production errors (*lapsus linguae*), which are fundamental in neurolinguistics for mapping cognitive processes in real-time; ii) Limited multimodality evaluation: although multimodal models were selected, the analysis protocol focused on text correction (unimodal). It did not explore how visual or auditory information might influence the interpretation of correction the contextual errors, which is relevant to the multimodal nature of human language; iii) Restriction of the zero-shot prompt: the use of the zero-shot prompt (direct instruction without prior examples) may have underestimate the true capability of the LLMs. Performance might have improved with few-shot learning or chain-of-thought prompting techniques, which are advanced methods for guiding the models' reasoning.

On the other hand, the study gives three possible future research directions: i) Expansion to oral production and dialectal errors: develop a corpus that includes spontaneous speech errors to compare human processing models (like Levelt's model) with LLM architecture, particularly regarding phonological errors and *lapsus linguae*. Also, investigate dialectal variations of Spanish (such as *seseo* or the aspiration of the [s]) and the LLMs' capacity to compensate for the "perceptual confusion" that these variations cause in humans; ii) Cognitive integration in LLM evaluation: apply evaluation metrics that focus not only on the final correction but also on the error explanation. Compare the quality of the explanations generated by LLMs with neurolinguistic taxonomies (like those for Broca's and Wernicke's aphasia) to see if the models "reason" about the affected processing level in a way that mimics human understanding; iii) The impact of pragmatic context: design experiments that test the LLMs' ability to detect errors that are only mistakes within a social or pragmatic context. This would involve testing the "understanding" of the intentionality behind an erroneous utterance.

Figure 1. Error detection rate of LLMs.

|         | Spelling Errors | Sintactic Errors | Lexical Errors | Semantic Errors | Verb 'haber' | Dequeísmo |
|---------|-----------------|------------------|----------------|-----------------|--------------|-----------|
| GPT-5   | 66.7%           | 92.7%            | 88,5%          | 76.8%           | 100%         | 100%      |
| GPT-4   | 61.4%           | 89.1%            | 81.2%          | 75.3%           | 100%         | 100%      |
| Grok    | 65.3%           | 76.5%            | 84.3%          | 81.2%           | 100%         | 100%      |
| Llama-3 | 54.7%           | 87.6%            | 78.8%          | 86.9%           | 100%         | 100%      |
| Gemini  | 53.5%           | 80%              | 75.6%          | 82.7%           | 100%         | 100%      |
| DeepSeek| 63.6%           | 83.4%            | 89.2%          | 79.7%           | 100%         | 100%      |

Figure 2. Detection Rates of Spanish Grammatical Errors by LLMs (%).

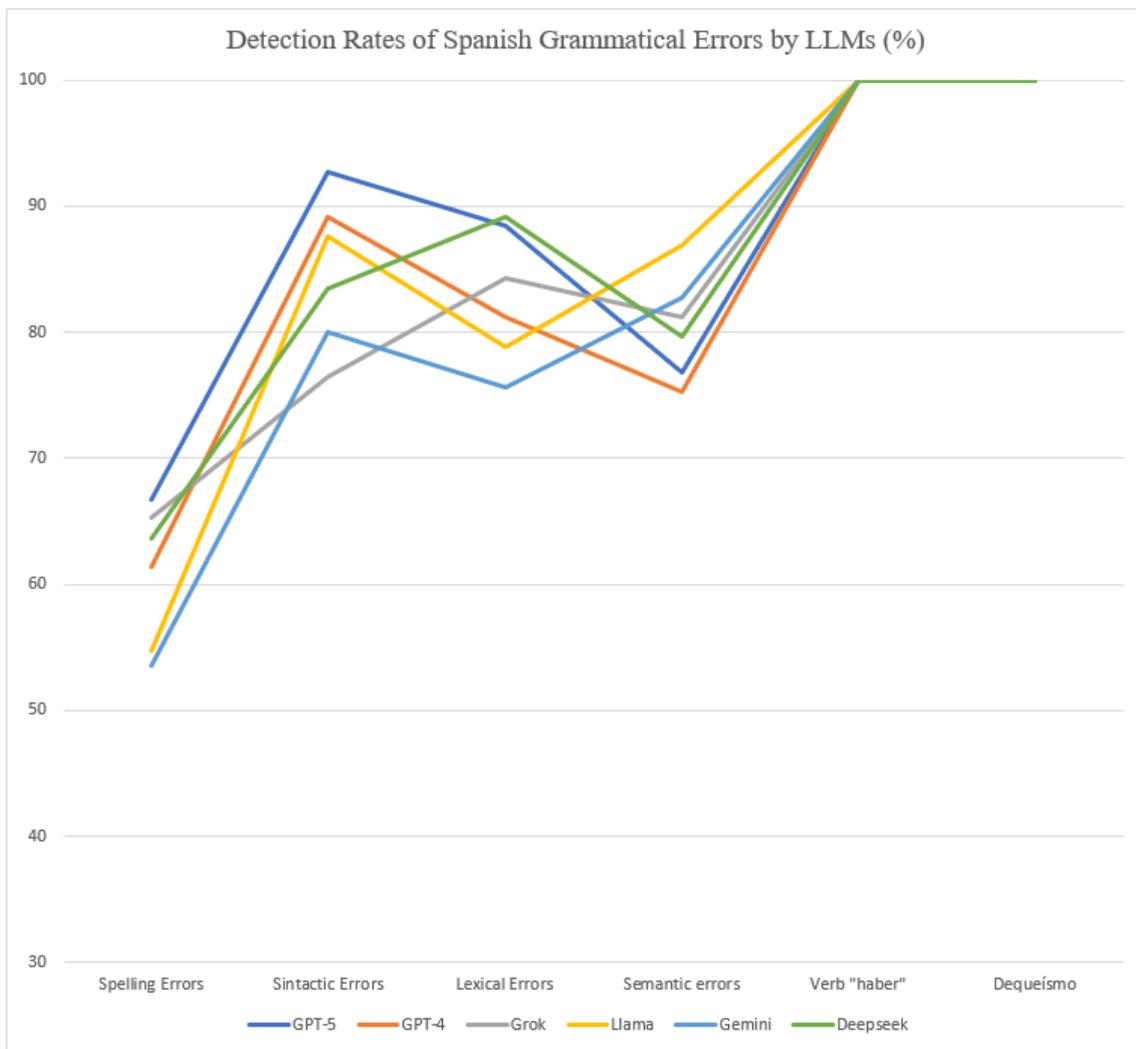

Figure 3. Summary of aphasias.

| Type of Aphasia | Damaged Brain Region | Characteristic Error | Processing Level Affected | Error Awareness |
|---|---|---|---|---|
| Broca's Aphasia | Left frontal lobe | Agrammatism, telegraphic speech | Syntactic-Morphological | High, often leading to frustration |
| Wernicke's Aphasia | Left temporal lobe | Paraphasias (literal and verbal), neologisms | Lexical-Semantic, comprehension | Low, patients usually unaware of errors |
| Anomic Aphasia | Variable, often temporal lobe | Anomia (difficulty retrieving words) | Lexical-Semantic | High |
| Global Aphasia | Extensive damage to language areas | Near-total loss of language | All levels | Low to none |